\RequirePackage{fix-cm}
\documentclass[smallextended]{svjour3}       
\smartqed  
\usepackage{graphicx}
\begin{document}

\title{Knowledge Graph for NLG in the context of conversational agents
}


\author{$\textrm{Hussam Ghanem}^\textrm{1}$ \and $\textrm{Massinissa Atmani}^\textrm{1}$ \and  $\textrm{Christophe Cruz}^\textrm{1}$  
}


\institute{
              ICB, UMR 6306, CNRS, Université de Bourgogne \\
              21000 Dijon, France \\
}

\date{Received: date / Accepted: date}

\maketitle

\begin{abstract}
The use of knowledge graphs (KGs) enhances the accuracy and comprehensiveness of the responses provided by a conversational agent. While generating answers during conversations consists in generating text from these KGs, it is still regarded as a challenging task that has gained significant attention in recent years. In this document, we provide a review of different architectures used for knowledge graph-to-text generation including: Graph Neural Networks, the Graph Transformer, and linearization with seq2seq models. We discuss the advantages and limitations of each architecture and conclude that the choice of architecture will depend on the specific requirements of the task at hand.
We also highlight the importance of considering constraints such as execution time and model validity, particularly in the context of conversational agents. Based on these constraints and the availability of labeled data for the domains of DAVI, we choose to use seq2seq Transformer-based models (PLMs) for the Knowledge Graph-to-Text Generation task. We aim to refine benchmark datasets of kg-to-text generation on PLMs and to explore the emotional and multilingual dimensions in our future work.
Overall, this review provides insights into the different approaches for knowledge graph-to-text generation and outlines future directions for research in this area.

\keywords{Conversational agents \and Knowledge graphs \and Natural Language Generation}
\end{abstract}

\section{Introduction}
\label{intro}
Conversational agents, also known as chatbots, are computer programs designed to simulate conversation with human users \cite{[171]}. These agents can be integrated with messaging platforms, mobile applications, and websites to provide instant support to customers and handle simple tasks, such as answering questions or helping with bookings.
Conversational agents using knowledge graphs (KG) \cite{[15]} are a type of chatbot that leverages structured data stored in a knowledge graph to generate human-like responses. The knowledge graph is a graph-based representation of entities and their relationships, providing a structured source of information for the chatbot to access and use during conversation. This enables the chatbot to provide more accurate and comprehensive answers to user's questions. The use of knowledge graphs can greatly enhance the capabilities of conversational agents and make interactions more informative and useful \cite{[173]}.\\

Generating answers during conversations consists in generating text from data. Data-to-text processes require algorithms that generate linguistically correct sentences for humans and express the semantics and structure of non-linguistic data (sequence, tree, graph, etc.). In addition, the generation of textual answers requires supporting several languages. And for a better interaction with a conversational agent, the emotional context of the conversation (Common Ground) is fundamental. The aim of this work in collaboration with the company DAVI is to integrate a socio-emotional dimension into human-machine interactions which complement the technical and “business” skills linked to professional expertise. The company DAVI is a software publisher in SaaS mode which has expertise in the fields of AI, Affective Computing, and Human Machine Interactions (HMIs). The following picture presents the composite AI of DAVI’s solution including Natural Language Understanding, Emotion detection, skills modeling, Natural Language Generation, and Body Language Generation.

\begin{figure}[ht]
\centering
  \includegraphics[width=0.99\textwidth]{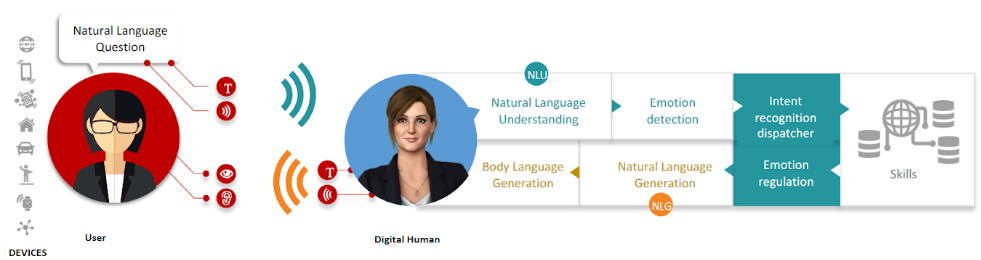}
\caption{Composite AI pipeline of virtual agents at DAVI}
\label{fig:1}       
\end{figure}

For now, the natural language generation (NLG) of a conversational engine does not benefit from the latest technological advances in natural language processing (NLP). The Natural Language processing step is based on a manual process to define the template of the answer. This process requires a costly amount of time. Thus, the purpose of this project is to automate and reduce the burden of the generation of template-based responses (as the responses are manually written through a set of rules) in the implementation of conversational agents. The template-based responses are modellized and stored in the skills' database. Regarding the emotional dimension, emotional responses are injected automatically in the answer depending the emotional analysis of the user. To automate NLG answers from Skills, knowledge graphs were identified.\\

\begin{figure}[ht]
\centering
  \includegraphics[width=0.7\textwidth]{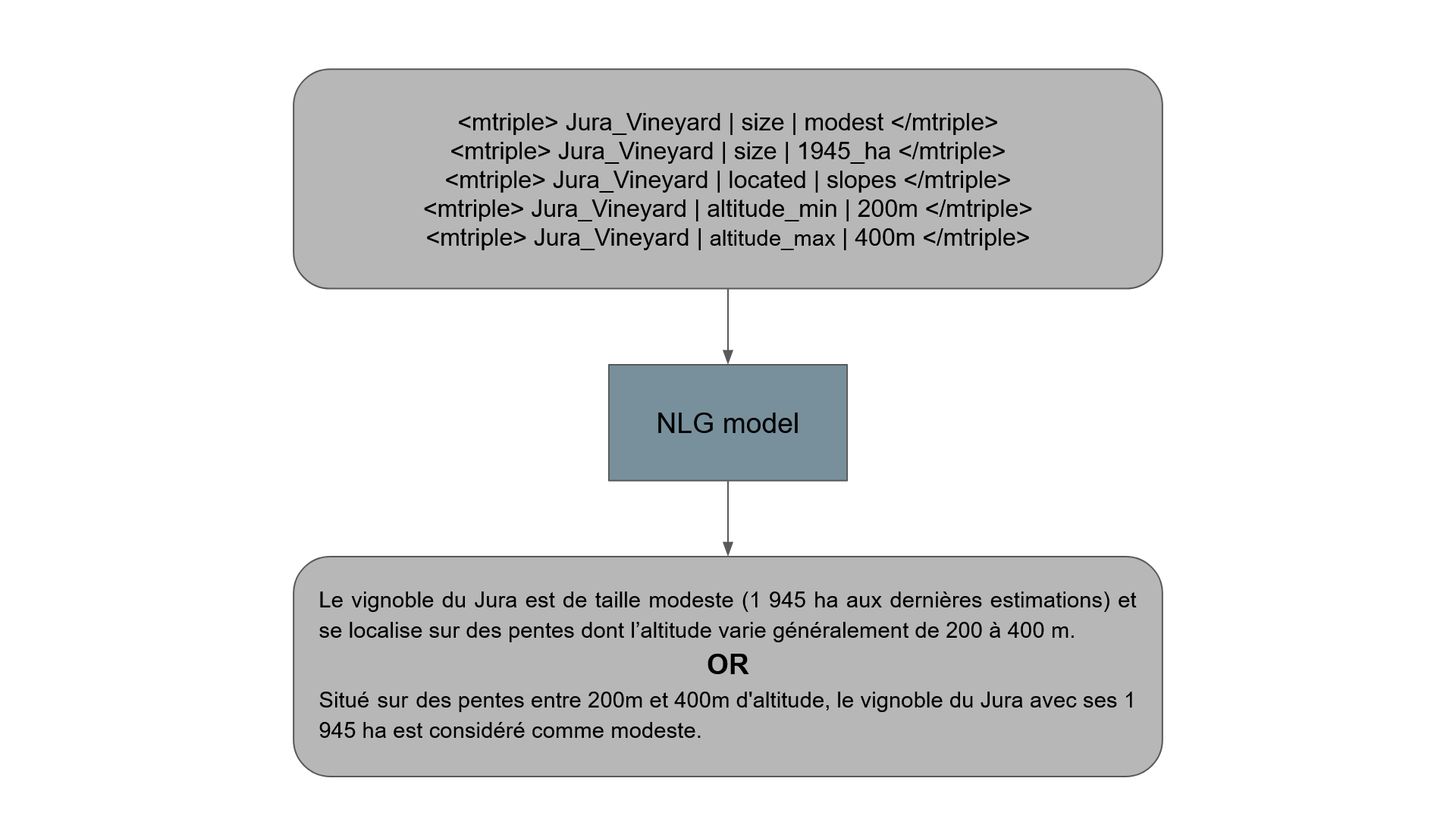}

\caption{ Automatic text generation from the knowledge graph}
\label{fig:2}       
\end{figure}

In NLG, Two criteria \cite{[94]} are used to assess the quality of the produced answers. The first criterion is semantic consistency (Semantic Fidelity) which quantifies the fidelity of the data produced against the input data. The most common indicators are 1/ \textbf{Hallucination}: It is manifested by the presence of information (facts) in the generated text that is not present in the input data; 2/ \textbf{Omission}: It is manifested by the omission of one of the pieces of information (facts) in the generated text; 3/ \textbf{Redundancy}: This is manifested by the repetition of information in the generated text; 4/ \textbf{Accuracy}: The lack of accuracy is manifested by the modification of information such as the inversion of the subject and the direct object complement in the generated text; 5/ \textbf{Ordering}: It occurs when the sequence of information is different from the input data. The second criterion is linguistic coherence (Output Fluency) to evaluate the fluidity of the text and the linguistic constructions of the generated text, the segmentation of the text into different sentences, the use of anaphoric pronouns to reference entities and to have linguistically correct sentences.\\

Today, neural approaches offer performances exceeding all classical methods for linguistic coherence. However, limits are still present to maintain semantic consistency, and their performance deteriorates even more on long texts \cite{[90]}. Another limitation due to the complexity of neural approaches is that text generation is non-parameterized with no control over the structure of the generated text. Thus, most of the current neural approaches arrive behind template-based approaches on the criterion of semantic consistency \cite{[91]}, but they are far superior to them on the criterion of linguistic consistency. This can be explained by the fact that large language models manage to capture certain syntactic and semantic properties of the language.\\

The present review is organized as follows, Section \ref{sec:1} presents a comprehensive overview of the current state-of-the-art approaches for knowledge graph-to-text generation. In Section \ref{sec:2}, we present the latest architectures and techniques that have been proposed in this field. Finally, Section \ref{sec:3} critically examines the strengths and limitations of these techniques in the context of conversational agents

\section{Knowledge Graph-to-Text Generation}
\label{sec:1}
KG-to-text generation aims at producing easy-to-understand sentences in natural language from knowledge graphs (KGs) while maintaining semantic consistency between the generated sentences and the KG triplets. Compared to the traditional text generation task (Seq2Seq), generating text from a knowledge graph is an additional challenge to guarantee the authenticity of the words in the generated sentences. The existing methods can be classified according to three categories (Figure 3) and will be detailed later:
\begin{itemize}
    \item \textbf{Linearisation with Sequence-to-sequence (Seq2Seq)}: convert the graph to a sequence which is the fed to a sequence-to-sequence model;
    \item \textbf{Graph Neural Networks (GNNs)}: encode topological structures of a graph and learn the representation of an entity by the aggregation of the features of the entities and neighbors. They are not used as a standalone model and require a decoder to complete the encoder-decoder architecture;
    \item \textbf{Graph Transformer (GT)}: the enhanced version of the original transformer adapted to handle graphs.
\end{itemize}

\begin{figure}[ht]
\centering
  \includegraphics[width=0.99\textwidth]{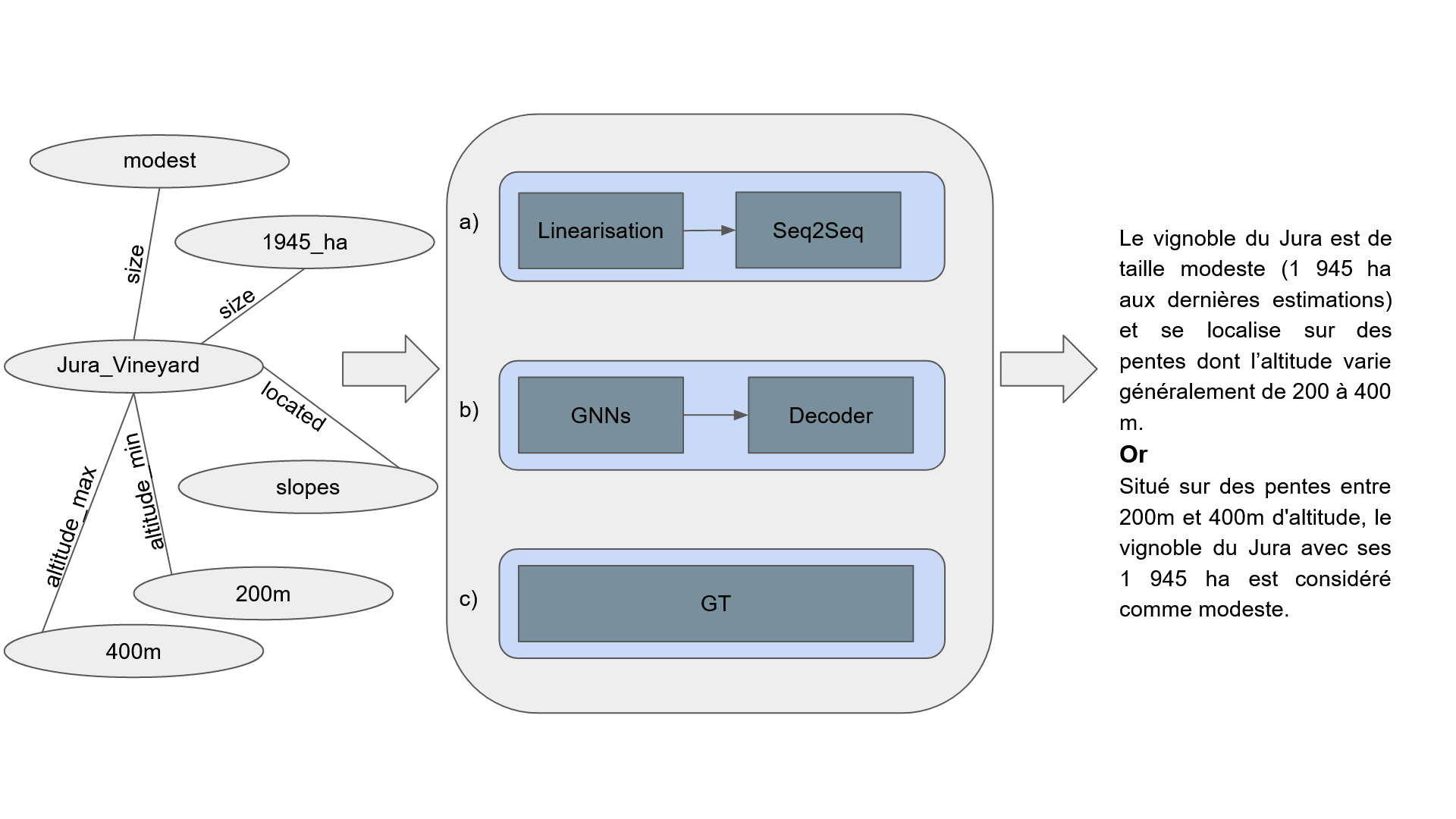}
\caption{The architecture of KG-to-text generation with the three categories of representation, a) Linearization + Seq2Seq, b) GNNs with decoder (e.g. LSTM), and  c) Graph Transformer (GT)}
\label{fig:3}       
\end{figure}

The term "knowledge graph" has been around since 1972, but its current definition can be traced back to Google's 2012 announcement of their Knowledge Graph. This was followed by similar announcements from companies such as Airbnb, Amazon, eBay, Facebook, IBM, LinkedIn, Microsoft, and Uber, among others, leading to an increase in the adoption of knowledge graphs by various industries \cite{[XX]}. As a result, academic research in this field has seen a surge in recent years, with an increasing number of scientific publications on knowledge graphs \cite{[XX]}. These graphs utilize a graph-based data model to effectively manage, integrate, and extract valuable insights from large and diverse datasets \cite{[YY]}.\\

Knowledge graphs, which are composed of nodes that represent different types of entities and edges that denote various types of relationships between those entities, are known as heterogeneous graphs. The integration of information from multiple sources and domains in knowledge graphs leads to an even greater degree of heterogeneity. To address this, recent research has applied heterogeneous graph embedding methods to represent knowledge graphs effectively. For example, ERNIE \cite{[104]} and KnowBERT \cite{[105]} employ knowledge graph embedding techniques such as TransE \cite{[106]} and TuckER \cite{[174]} to encode knowledge graphs.\\

Generating text and learning alignments between source entities/relationships and target tokens from scratch is a challenging task for standard language models because of the limited amount of parallel graph-text data \cite{[46],[97]}. To overcome this limitation, recent research has focused on developing generalist pre-trained language models for KG-to-text generation. A common approach is to linearize input graphs into text sequences and fine-tune pre-trained seq2seq Transformer models such as GPT \cite{[98]}, BART \cite{[16]}, or T5 \cite{[99]} based on KG-to-text datasets \cite{[1],[100]}. These pre-trained language models can generate high-quality texts with a simple fine-tuning to the target task, thanks to their self-supervised pre-training on large-scale corpora of unlabeled texts. In fact, pre-trained language models outperform other models with sophisticated structures in KG-to-text generation tasks. This type of approach will be detailed in section \ref{sec:2}.

According to \cite{[37]}, text generation tasks using KG-to-text models mainly fall under three aspects:
\begin{itemize}
    \item \textbf{Encoder modification}: To reduce the loss of structural information in sequence encoders with graph inputs that have been linearized \cite{[72],[8],[93]}, proposals concentrate on constructing more intricate encoder structures to improve the representation of graphs, including GNNs and GTs;
    \item \textbf{Unsupervised training}: These proposals consist in designing unsupervised training methods to jointly learn graph-to-text and text-to-graph conversion tasks with non-parallel graph-to-text data \cite{[103],[46],[48]}. This makes it possible to compare the final result of the process with the input data;
    \item \textbf{Build pre-trained models}: With the development of pre-trained Natural Language Generation (NLG) models such as GPT, BART, and T5, some recent work directly refines these models on graph-to-text datasets and reports significant performance \cite{[1],[100],[40],[3]}.
\end{itemize}

Compared to existing work on pre-trained models for KG-text generation, the JoinGT model \cite{[37]} uses pre-training methods to explicitly learn graph-text alignments instead of directly tuning the pre-trained models seq2seq on KG-to-text datasets.

\section{Architectures}
\label{sec:2}
In this section, the different architectures used in data-to-text tasks will be presented. As the nature of the data greatly influences the choice of the architecture of the neural approaches, most works either try to adapt the inputs to the architectures of the models or propose new architectures better adapted to the types of input data. Due to the nature of sentences and the tree structure of its representations, several works have proposed to model data structures of this type to enhance performance.

\subsection{Graph linearisation and sequence to sequence models (Seq2seq)}
\label{sec:2.1}

Recent years have been marked by significant achievements in the field of PLMs (Pretrained Language Models) \cite{[101],[116]}. Pre-trained on massive corpora, PLMs exhibit good generalization ability to solve related NLG (Natural Language Generation) downstream tasks \cite{[113]}. However, most existing PLMs were trained on textual data \cite{[116],[16]} without ingesting any structured data input. The seq-to-seq category consists of linearizing the KG \cite{[1],[75],[72],[4]} and then formulating a Seq2seq generation task using PLMs like GPT \cite{[98]}, BART \cite{[16]} or T5 \cite{[99]} with linearized KG nodes as input to generate sentences. The use of pre-trained language models (PLMs) in KG-to-text generation has shown superior performance, but still faces two major challenges: 1) loss of structural information during encoding, as existing models like BERT do not explicitly take into account the relationship between input entities; and 2) lack of explicit graph-text alignments, as complex knowledge graph structures make it difficult to learn graph-text alignments through text reconstruction-based pre-training tasks.
Despite attempts to retain as much of the graph topology as possible with seq2seq methods, the Transformer-based seq2seq models’ cost is not cheap (especially in the pretraining phase). Also, the computational cost of linearization can be high for large knowledge graphs. Hence, and so to better keep the graph topology, Graph Neural Networks (GNNs) have been proposed, which will be discussed in the next section.

\subsection{Graph Neural Networks (GNNs)}
\label{sec:2.2}

Different approaches use different variants of GNNs architectures such as GCNs (Graph Convolutional Networks) \cite{[82]} or extensions of GCNs such as Syn-GCNs \cite{[128]} or DCGCNs \cite{[17]}. Other approaches use the variant GATs (Graph Attention Networks) \cite{[11]}. Or approaches that use the GGNNs (Gated Graph Neural Networks) variant \cite{[42],[61],[137]}. Graph Neural Networks (GNNs) are a type of neural network that are well-suited for processing graph-structured data. In the context of knowledge graph-to-text generation, GNNs can be used to model the relationships between entities in a knowledge graph and generate text based on those relationships. Recent research on using GNNs for knowledge graph to text generation has shown promising results. Some studies have used graph convolutional networks (GCNs) \cite{[82]} to encode the relationships between entities in a knowledge graph into a low-dimensional representation, or extensions of GCNs such as Syn-GCNs \cite{[128]} and DCGCNs \cite{[17]}. Other studies have used graph attention networks (GATs) \cite{[11]} to dynamically weight the importance of different entities and relationships in the knowledge graph when generating text. Other studies have used a gating mechanism (Gated Graph Neural Networks or GGNNs) that allows for effectively controlling the flow of information between nodes in the graph, which is useful for incorporating contextual information \cite{[42],[61],[137]}. Additionally, some researchers have combined GNNs with reinforcement learning to generate text that maximizes a reward function defined over the generated text and the knowledge graph \cite{[61]} .\\

Overall, the use of GNNs for knowledge graph to text generation is an active area of research, with many recent studies exploring different architectures and training methods. The results so far suggest that GNNs can effectively capture the relationships between entities in a knowledge graph and generate high-quality text based on that information. A limitation relied to KG-to-Text generation with GNNs, is that GNNs can be computationally expensive and may struggle to handle large knowledge graphs. Additionally, their performance may degrade for graphs with complex relationships or structures. Despite these limitations, GNNs remain a promising direction for knowledge graph-to-text generation.

\subsection{Graphs Transformers (GTs)}
\label{sec:2.3}

In order to benefit from the power of models based on Transformer and to be able to model tree or graph-type data structures as with GNNs, and to overcome the limitations of local neighborhood aggregation while avoiding strict structural inductive biases, recent works have proposed to adapt the Transformer architecture. As Graph Transformers are equipped with self-attention mechanisms, they can capture global context information by applying these mechanisms to the nodes of the graph.\\

According to \cite{[18]}, GT differs from GNNs in that it allows direct modeling of dependencies between any pair of nodes regardless of their distance in the input graph. An undesirable consequence is that it essentially treats any graph as a fully connected graph, greatly reducing the explicit structure of the graph. To maintain a structure-aware view of the graph, their proposed model introduces an explicit relationship encoding and integrates it into the pairwise attention score computation as a dynamic parameter.\\

From the GNNs pipeline, if we make several parallel heads of neighborhood aggregation and replace the sum on the neighbors by the attention mechanism, e.g. a weighted sum, we would get the Graph Attention Network (GAT). Adding normalization and MLP feed-forward, we have a Graph Transformer \cite{[133]}. For the same reasons as Graph Transformer, \cite{[26]} presents the K-BERT model, they introduce four components to augment the Transformer architecture and to be able to handle a graph as input. The first knowledge layer component takes a sentence and a set of triplets as input and outputs the sentence tree by expanding the sentence entities with their corresponding triplets. They also add the Seeing Layer component to preserve the original sentence structure and model the relationships of the triples by building a Visibility Matrix. Another component is the Mask-Transformer where they modify the self-attention layer to consider the visibility matrix when calculating attention.\\

The use of Graph Transformers for Knowledge graph to text generation has gained popularity in recent years due to their ability to effectively handle graph structures and capture the relationships between nodes in the graph. Additionally, Graph Transformers can handle large graphs and have the ability to model long-range dependencies between nodes in the graph. Despite the advantages, the training of Graph Transformers can be computationally expensive and the interpretability of the model remains a challenge. Overall, the use of Graph Transformers for Knowledge graph to text generation is a promising area of research and can lead to significant improvements in the generation of text from knowledge graphs.

\section{Discusion}
\label{sec:3}

The Graph Neural Network (GNN), the Graph Transformer, and linearization with seq2seq models are three different architectures for knowledge graph to text generation, a task that involves generating natural language text from structured knowledge representations like knowledge graphs.
GNNs are a type of deep learning model that are well-suited for processing graph-structured data. They provide a flexible and scalable way to model the graph structure and relationships, but they may not be able to efficiently handle large and complex graphs.
The Graph Transformer is a specialized version of the Transformer, designed for graph-to-sequence learning tasks. It provides a more direct and efficient way to process the graph structure compared to linearization with seq2seq models, but it may require more training data and computation resources.
Linearization with seq2seq models is a simpler and easier to implement approach that involves converting the knowledge graph into a linear sequence, such as a sentence, and then using a seq2seq model to generate text from the linearized representation. However, this approach can lose some of the structural information and relationships in the knowledge graph during the linearization process.\\

In summary, each of the three architectures has its own advantages and disadvantages, and the choice of architecture will depend on the specific requirements of the actual task.\\

As our project is part of a context of conversational agents, we must take into account all the consequent constraints such as the execution time (to respect the instant conversation constraint) and the validity of the model where the answer must not be incorrect or ambiguous.\\
If most of the current models have satisfactory validity performances, the inference time of models based on GNN and GraphTransformer exceeds the limit threshold found in a fluid and natural conversation and requires a huge memory load that violates the standards of current industrialization (MLOps) of neural models.\\

In light of these elements, and with the constraint of the data labellisation for the domains of DAVI, we choose to go further with seq2seq Transformer based models (PLMs) in our Knowledge Graph-to-Text Generation. We also want to shed light on the fact that DAVI already handles such models in their pipeline and they have the knowledge and infrastructure to optimize the integration and deployment of the seq2seq models. Hence, DAVI should still remain in control of the time to market of the NLG solution.

\section{Conclusion}
\label{sec:4}
In conclusion, the document discusses different data-to-text architectures and highlights their advantages and limitations in the context of graph-to-text project with DAVI. The Graph Neural Network (GNN), Graph Transformer, and seq2seq models are three approaches that have been applied to the task of generating natural language text from structured knowledge representations like knowledge graphs. Each approach has its own advantages and disadvantages, and the choice of architecture will depend on the specific requirements of the task.
Considering the constraints of our project, which includes developing a conversational agent that must generate valid responses in real-time, we have decided to move forward with seq2seq Transformer-based models (PLMs) as they have satisfactory performance on validity and execution time. Additionally, DAVI already handles such models in their pipeline and can optimize their integration and deployment. Our next step will be to explore state-of-the-art approaches that take into account the emotional and multilingual dimensions to achieve the objectives of the graph-to-text project.

\end{document}